%% file: 00-main.tex
\documentclass[10pt,conference]{IEEEtran}

\usepackage{cite}
\usepackage{amsmath,amssymb,amsfonts}
\usepackage{soul}

\usepackage{graphicx}
\usepackage{textcomp}
\usepackage{xcolor}
\usepackage{booktabs}
\usepackage{ulem}
\usepackage[inline]{enumitem}
\usepackage{url}

\usepackage{amsmath}
\usepackage{algorithm}
\usepackage{algpseudocode}
\usepackage{balance}

\title{Fault Localization for Buggy Deep Learning Framework Conversions in Image Recognition\thanks{Authors, Ajitha Rajan and Nikolaos Louloudakis, would like to acknowledge support received from funding sources, UKRI Trustworthy Autonomous Systems Node in Governance and Regulation (EP/V026607/1) and Royal Society Industry Fellowship, for this work.}}

\author{\IEEEauthorblockN{Nikolaos Louloudakis}
\IEEEauthorblockA{\textit{n.louloudakis@ed.ac.uk} \\
\textit{University of Edinburgh\vspace{-3.5\baselineskip}}} \\
\and
\IEEEauthorblockN{Perry Gibson}
\IEEEauthorblockA{\textit{perry.gibson@glasgow.ac.uk} \\
\textit{University of Glasgow\vspace{-3.5\baselineskip}}} \\
\and
\IEEEauthorblockN{Jos\'e Cano}
\IEEEauthorblockA{\textit{jose.canoreyes@glasgow.ac.uk} \\
\textit{University of Glasgow\vspace{-3.5\baselineskip}}} \\
\and
\IEEEauthorblockN{Ajitha Rajan}
\IEEEauthorblockA{\textit{arajan@ed.ac.uk} \\
\textit{University of Edinburgh\vspace{-3.5\baselineskip}}} \\
}

\IEEEoverridecommandlockouts
\begin{document}

 \maketitle

%**********************************************************************************************

\begin{abstract}

When deploying Deep Neural Networks (DNNs), developers often convert models from one deep learning framework to another (e.g., TensorFlow to PyTorch).
However, this process is error-prone and can impact target model accuracy.
To identify the extent of such impact, we perform and briefly present a differential analysis against three DNNs widely used for image recognition (MobileNetV2, ResNet101, and InceptionV3) converted across four well-known deep learning frameworks (PyTorch, Keras, TensorFlow (TF), and TFLite), which revealed numerous model crashes and output label discrepancies of up to 100\%.
To mitigate such errors, we present a novel approach towards fault localization and repair of buggy deep learning framework conversions, focusing on pre-trained image recognition models.
Our technique consists of four stages of analysis: 
\begin{enumerate*}
    \item conversion tools,
    \item model parameters,
    \item model hyperparameters, and
    \item graph representation.
\end{enumerate*}
In addition, we propose various strategies towards fault repair of the faults detected.
We implement our technique on top of the Apache TVM deep learning compiler, and we test it by conducting a preliminary fault localization analysis for the conversion of InceptionV3 from TF to TFLite. Our approach detected a fault in a common DNN converter tool, which introduced precision errors in weights, reducing model accuracy. After our fault localization, we repaired the issue, reducing our conversion error to zero.

\end{abstract}

%**********************************************************************************************

\section{Introduction}
\input{01-Introduction}

\section{Related Work}
\input{02-Related_Work}

\section{Motivation}
\input{03-Motivation}
\label{section:motivation}

\section{Methodology}
\label{sec:method}
\input{04-Methodology}

\section{Conclusions and Future Work}
\input{99-Conclusions}

%\vfill \eject

\bibliographystyle{IEEEtran}
\balance
\bibliography{00-main.bib}

\end{document}

%% file: 01-Introduction.tex
Deep Neural Network (DNN) models, trained using a given deep learning (DL) framework (such as PyTorch~\cite{pytorch}, TensorFlow (TF)~\cite{tensorflow2015-whitepaper}), can be
converted to a different DL framework (such as Keras~\cite{cholletKeras2015}).
Common reasons for this conversion include  
\begin{enumerate*}
    \item deployment on resource-constrained environments such as IoT devices, which may require lightweight DL frameworks (e.g., TFLite), and
    \item support for a wider set of features, that allow more in-depth model modification and optimization, such as explicitly defining forward propagation implementation.
\end{enumerate*}
Conversion of DNN models between DL frameworks is facilitated by automated conversion processes using tools such as \texttt{tf2onnx}~\cite{tf2onnx}, \texttt{onnx2keras}~\cite{onnx2keras}, \texttt{onnx2torch}~\cite{onnx2torch}, and \texttt{MMdnn}~\cite{liuEnhancingInteroperabilityDeep2020}. 
However, this conversion process can introduce
faults~\cite{louloudakis2022assessing, collie2020m3, yaneva2017compiler,TF-errors}, which
can make the converted models undeployable or reduce performance on their target task~\cite{chen2020comprehensive,chen2021empirical}.

In order to mitigate this problem, we propose an automated approach for fault localization and repair of faults introduced by the DL framework conversion process. 
We focus on DL framework conversion used in deployment of pre-trained image recognition models, utilized for image classification tasks.
Note that our methodology is agnostic to the DNN architecture and can be applied to other tasks such as image segmentation.
Our approach detects faults introduced in model parameters, hyperparameters, and the model graph, as the primary coefficients that define DNN model behaviour.
The proposed approach performs analysis and comparison against source and target model parameters and hyperparameters, as well as comparison of layer activations for inputs resulting in output label discrepancies against the source and the target model.
Additionally, we explore potential discrepancies introduced by graph transformations between the source and the target model during the conversion process.
Then, we propose a set of strategies to mitigate conversion faults such as the replacement of model parameters of the target model with those from source, and applying graph transformations that eliminate the error from the converted model.
Finally, we present an evaluation example of the conversion process for the InceptionV3 model converted from TensorFlow to TFLite. 
Our technique is able to detect precision errors in weights related to convolutional layers introduced by the TFLiteConverter tool, with value deviations of up to $0.01$ between \textit{Source} and \textit{Target}, which, although small, affected the model performance.

% Overall, in this paper we make the following main contributions:
Overall, the main contributions of this paper are:
\begin{enumerate*}
    \item A novel method to systematically localize faults in DL framework conversion processes, and         
    \item repair strategies for said faults.
\end{enumerate*}

%% file: 02-Related_Work.tex
A number of studies have been conducted related to faults introduced in the deployment process of DNNs. 
For instance, a study of 3023 Stack Overflow posts built a taxonomy of faults and highlighted the difficulty of DNN deployment~\cite{chen2020comprehensive}.
Another study explores the effect of DNN faults on mobile devices by identifying 304 faults from GitHub and Stack Overflow~\cite{chen2021empirical}, while other studies provide surveys on existing contributions towards machine learning testing components, workflow and application scenarios~\cite{zhang2019machine}. In addition, there are works related to exploring the test oracle problem in the context of machine learning~\cite{fontes2021using, tsimpourlas2021supervised}.
In terms of fault localization, DeepCover~\cite{sun2019deepcover} attempts to apply a statistical fault localization approach, focusing on the extraction of heatmap explanations from DNN inputs.
DeepFault~\cite{deepfault} focuses on a suspiciousness-oriented spectrum analysis algorithm in order to detect parts of the DNN that can be responsible for faults, while it also proposes a method for adversarial input generation. 
DeepLocalize~\cite{wardat2021deeplocalize} attempts to detect faults in DNNs by converting them to an imperative representation and then performing dynamic analysis on top of its execution traces.
Regarding fault localization in DL framework specifically, CRADLE~\cite{pham2019cradle} tries to detect faults introduced by DL frameworks by performing model execution graph analysis.
LEMON~\cite{wang2020lemon} leverages the metrics used by CRADLE for its analysis to apply mutation testing.

Although the above works attempt to overcome fault localization challenges for DNNs, none of them considers model conversions as a factor of fault introduction in DNNs and, therefore, no previous work explores this problem.
However, several tools exist to ease the DL model conversion process, including MMdnn~\cite{liuEnhancingInteroperabilityDeep2020}, tf2onnx~\cite{tf2onnx}, onnx2keras~\cite{onnx2keras}, onnx2torch~\cite{onnx2torch}, and tflite2onnx~\cite{tflite2onnx}.
There are also some native APIs for DL framework conversion to ONNX found within PyTorch~\cite{pytorch} and TFLite~\cite{tensorflow2015-whitepaper}.
These tools are extensively used, as they all have more than 100 stars on their GitHub repositories.
In addition, a recent study by Openja et al.~\cite{dlfconversionsstudy} highlights the challenges of the conversion process, while our preliminary work~\cite{louloudakis2022assessing, louloudakis2022exploring} explores the robustness of DNNs against different computational environment aspects, including DL framework conversions. However, the impact of DL framework conversions on DNN model correctness is not explored in-depth in the literature.

To the best of our knowledge, this paper is the first attempt focusing on the error proneness, fault localization, and repair of DL framework conversions for DNN models.
We focus on image recognition models as a a starting point, but our work is applicable to DNNs used in other domains.

%% file: 03-Motivation.tex
To observe the potential impact of DNN model conversions, we conducted an initial evaluation using three widely used image recognition models of varying size and architectural complexity: MobileNetV2~\cite{mobilenetv2}, ResNet101~\cite{resnet}, and InceptionV3~\cite{inceptionv3}. 
For each model, we used pre-trained versions from official repositories of four different DL frameworks: TensorFlow~\cite{tensorflow2015-whitepaper}, TFLite~\cite{tensorflow2015-whitepaper}, Keras~\cite{chollet2015keras}, and PyTorch~\cite{pytorch}.
We refer to the pre-trained model of each DL framework as the \textit{Source} model. 
As a result, we have 4 \textit{Source} versions for each of our 3 models.
We then convert each \textit{Source} model to use a different DL framework; we refer to the converted model as \textit{Target}. 
To implement the conversion, we use tools that convert the \textit{Source} model either directly to \textit{Target}, or to the ONNX~\cite{onnxsite} format, a popular model representation format that is designed to act as a common interchange format between frameworks.
Some DL frameworks, such as PyTorch and TFLite, have native tools for this conversion; whereas for others, such as TensorFlow, we leverage popular third-party conversion tools like tf2onnx~\cite{tf2onnx}.
We then convert from ONNX to \textit{Target} using a number of widely used libraries, such as onnx2keras\cite{onnx2keras} and onnx2torch~\cite{onnx2torch}.
Following the conversion process, we perform pairwise comparison between \textit{Source} and \textit{Target} model inferences using the ILSVRC 2017 object detection test dataset~\cite{ILSVRC17}, in order to detect discrepancies in classification introduced by the conversion process.

\begin{figure}
     \centering
     %\vspace{-32pt}
           % \advance\leftskip-1.2cm
     \includegraphics[width=0.99\columnwidth]{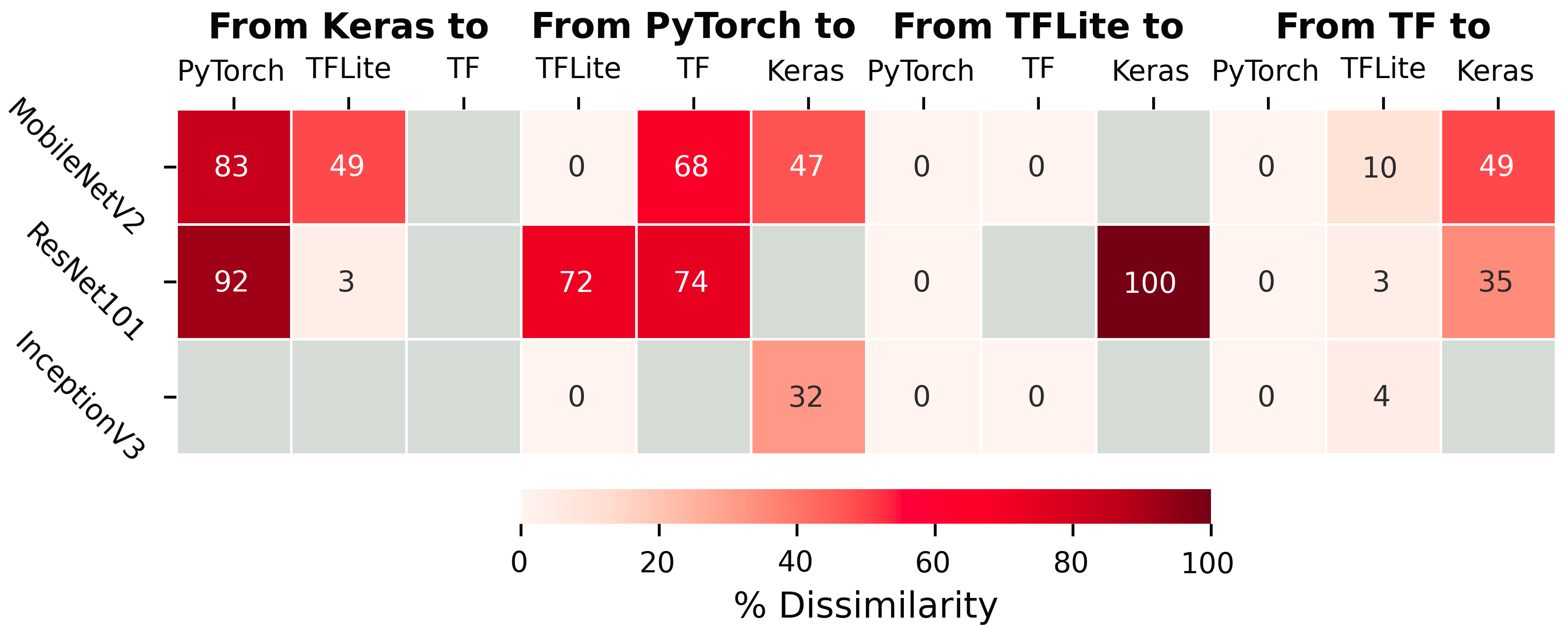}
     %\vspace{-20pt}
     \caption{Pairwise comparison of output labels between \textit{Source} and converted \textit{Target} models.}
     \label{fig:conversionsorig}
     %\vspace{-12pt}
\end{figure}

For each image of the dataset, we compare the output labels of \textit{Target} against \textit{Source} to check if any errors were introduced by the model conversion. 
The proportion of output label dissimilarities between \textit{Source, Target} pairs across all images in the dataset is shown in Figure~\ref{fig:conversionsorig}. 
As can be seen from the empty grey cells, the conversion tool crashes in 11 out of the 36 conversions across the three DNN models, indicating that the conversion process failed. 
This happened due to compatibility issues between the conversion tool and a given model architecture, or the \textit{Source} or \textit{Target} DL framework. 
Additionally, we observe 15 cases where the conversion succeeded without crashing, but the \textit{Target} model presented label discrepancies in comparison to the \textit{Source} model, with a maximum observed discrepancy of 100\% in the output labels when converting the ResNet101 model from TFLite to Keras.
The conversion of Keras to TFLite gives varying results across models, with MobileNetV2 and ResNet101  having considerable amounts of dissimilarity (49\% and 3\%, respectively), and InceptionV3 leading to a crash.
This points to weaknesses in the conversion tool with certain model architectures. 
Finally, for conversions between TF or TFLite to PyTorch no conversion errors were observed, while when converting TF to TFLite across all models we see relatively small discrepancies, 3-10$\%$, demonstrating a more reliable conversion. 
However, even small discrepancies may have non-negligible impact when these models are used in safety critical applications.

From Figure~\ref{fig:conversionsorig}, it is clear that the conversion process is error-prone and there is a need for a technique to localize and fix faults introduced by DL frameworks converters. 
We discuss our approach for fault localization and repair in Section~\ref{sec:method}.

%% file: 04-Methodology.tex
\begin{figure}[!t]
 \centering
 \includegraphics[width=0.99\columnwidth]{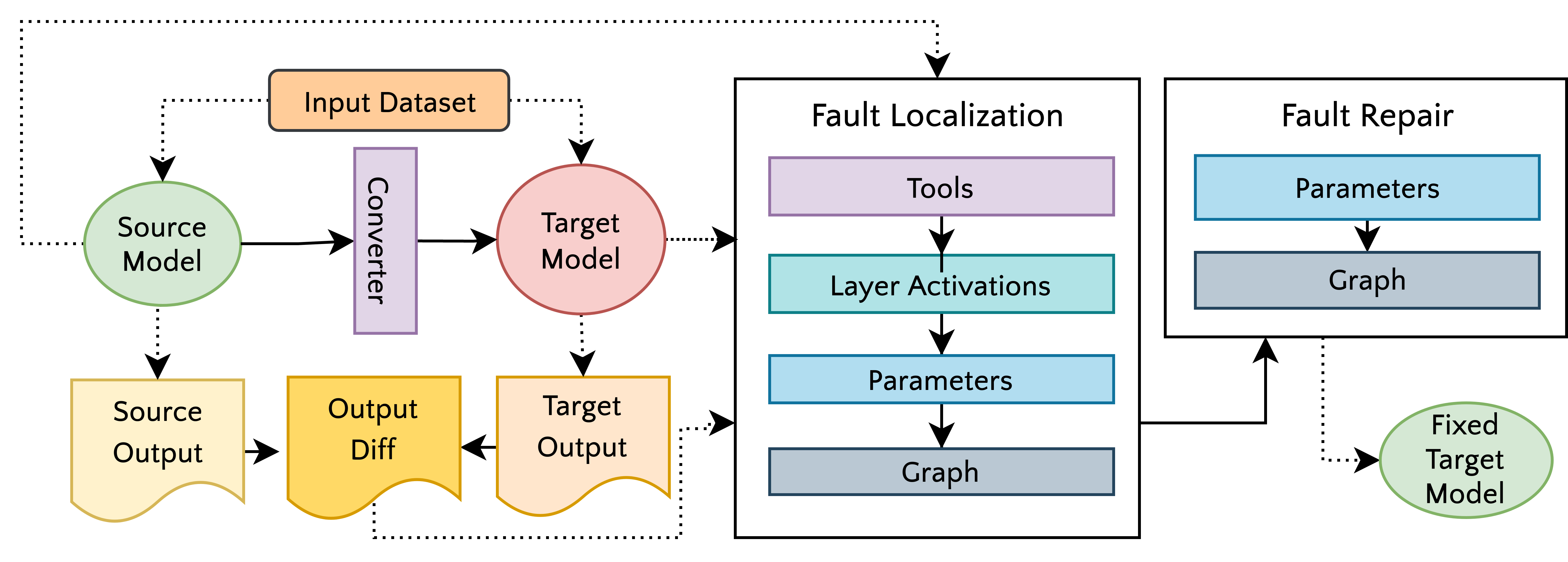}
 \caption{Fault Localization \& Repair Pipeline.}
 \label{fig:fault_localization}
 %\vspace{-12pt}
\end{figure}

The stages in our proposed approach for fault localization and repair are shown in Figure~\ref{fig:fault_localization}. 
It starts by converting a given model from \textit{Source} to \textit{Target} DL framework, then performs inference over both models over an input dataset, compares output labels to identify parts of the dataset which led to different outputs, and finally performs a fault localization and repair process where possible/appropriate.
We describe the fault localization and repair steps below.

%**********************************************************************************************

\subsection{Fault Localization \& Repair}

We start by examining the tools involved in the DNN model conversion to identify if the fault is introduced during conversion from \textit{Source} to the ONNX format, or from ONNX to \textit{Target}.  
We then complement this analysis by examining differences for three key DNN model architecture aspects:
\begin{enumerate*}
    \item Hyperparameters (such as kernel and batch size),
    \item parameters (such as weights and biases), and
    \item the model's graph structure (such as operations and their connections).
\end{enumerate*}
We describe these steps below.

%**********************************************************************************************

\subsubsection{Conversion Tools Analysis}

Following the DNN model conversion process, when discrepancies are observed between the \textit{Source} and the \textit{Target} model, it is important to identify which part of the conversion process was responsible. 
The conversion process typically uses more than one tool, e.g., one for conversion to ONNX format from \textit{Source}, and another to convert from ONNX to \textit{Target}. 
We explore this over a subset of the images that presented discrepancies between \textit{Source} and \textit{Target} while also considering the intermediate ONNX representation.
In particular, we consider a subset of the dataset inputs that presented different outputs between \textit{Source} and \textit{Target} models. 
In addition, we  perform inference using the ONNX intermediate representation from the conversion process, and compare the outputs against the \textit{Source} and \textit{Target}.
If the conversion process involves multiple steps, we repeat the process for all intermediate steps, so that we can better localize where the fault is introduced.

%**********************************************************************************************

\subsubsection{Parameter Analysis}

A correct DNN model conversion should result in a target model having the same parameters, and producing the same output as the source model.
However, if for some reason the parameters are altered (e.g., due to a precision error in the conversion process), this could potentially affect the model's correctness.

To detect this fault, we take the \textit{Source} and \textit{Target} model variants, and extract their parameters (e.g., weights and biases).
We then compare the parameters between model variants across layers of the same type (e.g., convolutions, bias additions, etc), by computing $\operatorname{mean}(\operatorname{abs}(P_{\mathit{source}} - P_{\mathit{target}}))$, where $P_{\mathit{source}}$, $P_{\mathit{target}}$ are the parameters of the source and converted target models, respectively. 
The value of the mean difference is expected to be zero when the model parameters are unaffected, and any other value indicates that there is a difference across the parameters in a specific layer, which is a potential cause for bugs.

%----------------------------------------

\subsubsection{Hyperparameter Analysis}

Much like parameters, incorrectly converted hyperparameters are another potential source of error.
For example, we would expect for a convolutional layer, the padding, strides, dilation, and other configurations would remain unchanged during a conversion.
However, a difference could indicate a potential source of error and is marked for further evaluation in our fault localization approach.is 

%----------------------------------------

\subsubsection{Layer Analysis}

To detect faults that occur on a specific layer, we propose a layer-based dynamic analysis approach.
Using a small, indicative subset of 5 images from the dataset that present output label discrepancies between \textit{Source} and \textit{Target} models, we perform inference and compare per-layer activations between the models. 
For each input, we compute the mean of differences found across activations for each layer.
We then further examine the layers affected sequentially, starting from the first layer and moving forward.
We focus on errors in the graph representation of that layer, as well as on implementation details. 
In particular, we examine if a layer or its graph neighbors are implemented in a different manner or are using different but equivalent operations (e.g., \textit{reshape} and \textit{flatten}) between the \textit{Source} and \textit{Target} model.

%----------------------------------------

\subsubsection{Fault Repair}

Once a difference is detected, we attempt one of the following options based on the location of the difference for fault repair.

(a) For differences in model hyperparameters, weights, and biases, the respective values from source can be replaced by the target model.
Since the conversion process should preserve those values, then the replacement in the target model should resolve the observed differences.

(b) For differences detected in layer activations, there are a number of measures that can be applied. 
First, a set of mappings can be applied in order to perform in-place replacement of parts of the graph that should behave similarly, but differences in implementation (such as the selection of a different layer type, or the addition of extra redundant layers) could cause differences in layer outputs.
For example, we observed cases (e.g., MobileNetV2, PyTorch-to-Keras conversion) where the \textit{flatten} layer was replaced by a \textit{reshape} layer by the converter tools. 
We instruct a layer replacement to the target based on the layer of the source model, while adjusting tensor inputs and outputs to preserve model validity.
In addition, if there are extra nodes added close to the layer affected, they could be modified and removed as an attempt to eliminate errors.
For instance, we observed the addition of some padding layers to the target model for a number of conversions (e.g., MobileNetV2, TF-to-PyTorch conversion). 
A potential fix is to simply remove this node.

Our current approach has limitations for cases where whole sub-graphs in the \textit{Target} model have completely different structure than the \textit{Source}. 
A replacement in this scenario is non-trivial and is subject to consideration for future work.
Once a fix is applied, inference is performed with the target model against the inputs causing discrepancies, and the behavior is monitored. 
If an improved result is detected for some or all of the images, then the fix is considered successful.

%**********************************************************************************************

\subsection{Implementation Details}

Our methodology is implemented using Apache TVM~\cite{tvm}, a cross-platform machine learning compiler framework.
We use TVM in order to build and perform inference for the \textit{Source} and \textit{Target} models, while we extract the graph parameters, graph structure provided by each model for weights, biases, and hyperparameters utilizing the model static parameters and graph description metadata generated across the build process.
We also use ONNXRuntime~\cite{onnxruntime} to perform intermediate representation inference.
In addition, we utilize the TVM Debugger to extract layer activations upon inference, as well as set specific inputs and extracting targeted outputs from hidden layers.
The TVM debugger was also used in order to achieve model repair strategies, such as replacing weights, biases, and hyperparameters. 
For the graph modification part, we utilized the ONNX~\cite{onnxsite} API in combination with ONNXModifier~\cite{onnx-modifier} in order to apply graph modifications.
We also used Netron~\cite{netronapp} for DNN graph observation purposes.

%**********************************************************************************************

\subsection{Preliminary Evaluation}

As an initial case study, we consider the conversion of InceptionV3 using TensorFlow (TF) as \textit{Source} and converting it to TFLite as \textit{Target}. 
The conversion involved two utilities, the native API of \textit{TFLite} (\textit{TFLiteConverter}), and \textit{tf2onnx}.
We observed label differences between \textit{Source} and \textit{Target} models for 4\% of the input images (240 out of 5500 images).
We were interested in this particular case study because the conversion error is low but still present in a small number of images.
This ``subtle'' failure is of particular interest for safety critical applications, where edge case behavior is more important.

For fault localization, we start by performing an analysis of the conversion tools on the images showing label differences. 
As seen in Table~\ref{tab:comparisonsconversions}, for three sample images label differences occur when converting 
models from TF to TFLite, but not in the conversion to ONNX. 
As a result, we find that the TFLiteConverter is the problematic part of this particular conversion process.
We perform further investigation of this tool in the next steps.

\begin{table}[!ht]
\centering
\small
 % \vspace{-10pt}
    \caption{Top-1 inference of images for InceptionV3 using TF, intermediate ONNX and TFLite converted from TF. }
    \label{tab:comparisonsconversions}
    \begin{tabular}{llll}
    \toprule
        Image ID & TF  & TFLite (TF) & ONNX \\ \midrule
        Image 1 & drum & drum & drum \\ \midrule
        Image 2 & wallet & purse & purse \\ \midrule
        Image 3 & wallaby & It. greyhound & It. greyhound \\ \bottomrule
    \end{tabular}
\end{table}

\begin{figure}[!t]
 \centering
 %\vspace{-10pt}
 \advance\leftskip-0.2cm
 \includegraphics[width=0.99\columnwidth]{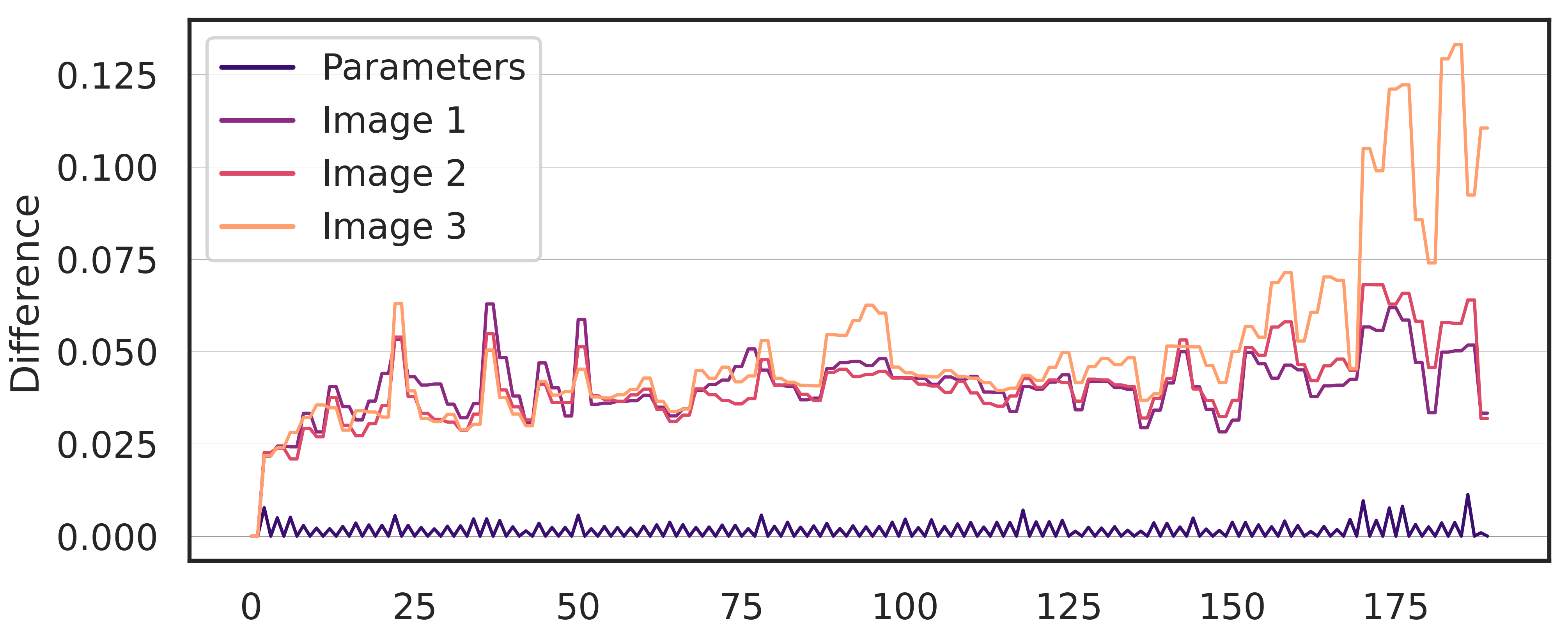}
 \caption{
 Layer-wise evaluation of the differences between InceptionV3 model sourced from TensorFlow, and converted to TFLite.
 \textit{Parameters} shows the mean difference between their weights and biases for convolutional and bias addition layers. 
 \textit{Image 1}, \textit{Image 2}, and \textit{Image 3} show models' differences in activations for two inputs across \textit{Source} and \textit{Target} models.}
 \label{fig:layers-mean}
\end{figure}

We then proceed with \textit{parameters} and \textit{layers} analysis between \textit{Source} and \textit{Target} to further examine the effects and the potential reasons for the problem.
% to the converted \textit{Target} model.
We consider an image presenting no discrepancies and two images presenting minor and major label discrepancies (by calculating and comparing Kendall's Tau coefficient~\cite{kendall} for the top-5 inference labels).
We present the results in Figure~\ref{fig:layers-mean}, where the \textit{Parameters} line indicates the mean of differences per-layer (x-axis) in parameters for two types of layers, convolutions and bias additions.
The remaining lines depict the differences in activations (mean of tensor values comparison) for each layer between \textit{Source} and \textit{Target}.
Image 1 presented no discrepancies, Image 2 presented small discrepancies, and Image 3 presented major discrepancies, measured using Kendall's Tau coefficient.
We observe layer 2 started presenting discrepancies between \textit{Source} and \textit{Target} for all images under test, affecting the model early in the process. Additionally,  there is a spike in the difference observed in layers 170 onwards for Image 3 (which presented large discrepancies between \textit{Source} and \textit{Target}).
We examined if the cause of the discrepancy was errors introduced in the model weights while using TFLiteConverter in the conversion process.
In particular, we performed a manual \textit{Source} and \textit{Target} model parameters inspection using Netron~\cite{netronapp}, which confirmed the fault localization finding, as we observed precision errors in the generated ONNX graph from \textit{Source}. 

In order to fix the error, we replaced the model weights of the \textit{Target} model with those from the \textit{Source}, and performed inference against the subset of images presenting discrepancies between the models. 
The outputs of the updated \textit{Target} were identical to the original \textit{Source}, resolving the issue, and proving its cause.

%% file: 99-Conclusions.tex
We presented a novel fault localization approach for errors encountered during DL framework conversion in image recognition models.
\iffalse
The need of such a technique is supported by the facts that
\begin{enumerate*}
    \item the conversion process is well-established amongst DNN developers and researchers, proven by the existence and the wide usage of related tools,
    \item our findings in preliminary experiments that indicate the extent of the error proneness of the procedure, and
    \item the lack of research in this particular area.
\end{enumerate*}
\fi 
It focuses on key DNN model elements such as parameters, hyperparameters, and graph architecture. 
We also propose strategies to repair the detected errors, such as correcting corrupted model weights.
As an example, we examined InceptionV3 when converted from TF to TFLite, which resulted in discrepancies for a small fraction of the input images. 
We used our approach to localize the conversion bug and fix it.
As future work, we aim to evaluate our approach against all conversion tools in Figure~\ref{fig:conversionsorig} and other image recognition models.
We will also apply it to other DL tasks such as object detection.
Finally, we plan to expand our fault repair strategies to address conversion errors that cause significant changes in \textit{Source} and \textit{Target} model graphs.